\documentclass[10pt,twocolumn,letterpaper]{article}

\usepackage[pagenumbers]{arxiv}

\usepackage{graphicx}
\usepackage{amsmath}
\usepackage{amssymb}
\usepackage{booktabs}
\usepackage[inline]{enumitem}
\usepackage[breaklinks,colorlinks]{hyperref}
\usepackage{hyphenat}
\usepackage[capitalize]{cleveref}

\begin{document}

\title{Face Detection on Mobile: Five Implementations and Analysis}

\author{Kostiantyn Khabarlak\\
Dnipro University of Technology\\
Ukraine\\
{\tt\small habarlack@gmail.com}
}
\maketitle

\begin{abstract}
  In many practical cases face detection on smartphones or other highly portable devices is a necessity. Applications include mobile face access control systems, driver status tracking, emotion recognition, etc. Mobile devices have limited processing power and should have long-enough battery life even with face detection application running. Thus, striking the right balance between algorithm quality and complexity is crucial. In this work we adapt 5 algorithms to mobile. These algorithms are based on handcrafted or neural-network-based features and include: Viola-Jones (Haar cascade), LBP, HOG, MTCNN, BlazeFace. We analyze inference time of these algorithms on different devices with different input image resolutions. We provide guidance, which algorithms are the best fit for mobile face access control systems and potentially other mobile applications. Interestingly, we note that cascaded algorithms perform faster on scenes without faces, while BlazeFace is slower on empty scenes. Exploiting this behavior might be useful in practice.
\end{abstract}


\blfootnote{
  Citation: \nohyphens{K.~Khabarlak ``Face Detection on Mobile: Five Implementations and Analysis,'' \emph{CoRR}, vol. abs/2205.05572, 2022. arXiv: \href{https://arxiv.org/abs/2205.05572}{2205.05572}.}
}

\section{Introduction}\label{sec:intro}

Low face detection inference time is crucial for mobile applications, such as driver sleepiness tracking, emotion recognition or mobile face access controls systems~\cite{FastLandmark,RfidAccessControl}. However, many of the recent research papers focus on algorithm quality, disregarding inference time. This makes many of the novel high-quality algorithms inapplicable on mobile.

In this work we provide some guidance on the algorithms best applicable to smartphones and other low-power portable devices. We have designed a special Android testing application, where we implement and analyze performance of 5 algorithms on mobile devices. The algorithms include well-known algorithms based on handcrafted features, such as Viola-Jones (Haar cascade), LBP, HOG, and more recent neural-network-based approaches, like MTCNN, BlazeFace. The analysis includes tests on low- and high-end devices, at different input image resolutions and on scenes with different number of faces. Interestingly, we find that inference time of algorithms typically depends on number of faces on the image. Based on the analysis conducted, recommendations on which algorithms should be preferred are provided.

\section{Algorithms Under Consideration}

We start with older, more established face detection methods. These algorithms are based on handcrafted features and a classifier (commonly, tree-based) that decides, which patch of an image contains a face. Only the classifier and not feature extraction logic is learned during training process. The first method is Haar cascade classifier, also known as Viola-Jones face detector~\cite{ViolaJones}. At the time of its inception it proposed one of the best quality to inference time ratios. To perform face detection, the image is split into overlapping sub-windows. Haar features combined with AdaBoost classifier are used to perform classification on each of the windows. Each window is then assigned label ``definitely not a face'' or ``possibly a face''. This allows to quickly reject regions that do not contain faces. Regions that possibly contain a face are then fed into the next detection stage. Thus, forming a cascade. Most face regions are rejected at early stages, which makes the algorithm fast. In addition, the authors have proposed a faster way to compute Haar filters for the whole picture by constructing the so-called integral image.

Other well-known face detection algorithms include Local Binary Pattern (LBP) method. The algorithm follows the idea of Viola and Jones, but uses LBP features that are faster to compute. Both Haar cascade and LBP algorithms are available in OpenCV~\cite{OpenCV} computer vision library. Another face detection method is based on Histogram of Oriented Gradients (HOG). First introduced in~\cite{HOG} for human detection and later adapted to face detection. HOG features are combined with SVM classifier. This algorithm is implemented in Dlib library~\cite{Dlib}.

In contrast to the above-mentioned methods, recent neural-network-based algorithms do not split the image into-subregions. The whole image is processed at once, which is more efficient and enables prediction of bounding boxes of non-standard sizes. The problem is then framed as combined regression and classification. To get coordinates of face bounding boxes regression is used, while classification predicts probability that certain bounding box contains a face. We have selected MTCNN~\cite{MTCNN} and BlazeFace~\cite{BlazeFace} algorithms for this section. We have adapted PyTorch~3.9~\cite{PyTorch} implementations of these algorithms to a smartphone.

Multi-task Cascaded Convolutional Neural Networks (MTCNN)~\cite{MTCNN} is the first of the selected algorithms, that is based on neural networks. According to authors, better face detection accuracy is achieved when solving several tasks at the same time:
\begin{enumerate*}[label={\arabic*)}]
  \item face bounding box regression;
  \item probability prediction that each bounding box contains a face;
  \item face landmark detection (location of eyes, nose tip, mouth corners).
\end{enumerate*}
MTCNN has several networks in a cascade:
\begin{enumerate*}[label={\arabic*)}]
  \item P-Net is a fast network that processes the image at multiple resolutions and predicts initial face bounding boxes;
  \item R-Net refines the predictions;
  \item O-Net further refines the predictions and outputs final bounding boxes.
\end{enumerate*}
To filter out overlapping bounding boxes Non-Maximum Suppression algorithm is used. Note, that face landmark detection performed in this algorithm may be used in many applications~\cite{FastLandmark}.

BlazeFace~\cite{BlazeFace} is a novel face detection algorithm. The only algorithm in this work, which primarily targets smartphones. The authors base on Single-Shot Detector (SSD)~\cite{SSD} with MobileNetV2~\cite{MobileNetV2} backbone. Despite the fact, that MobileNetV2 is a mobile-friendly neural network, it can still be improved. Bottleneck block is the key building block of the MobileNetV2 architecture. The authors propose to enlarge block's receptive field by using convolutions with larger kernel sizes. At the same time they reduce the overall number of Bottleneck blocks. This results in smaller inference time at comparable quality. The next step was to simplify SSD architecture to become a better fit for face detection. SSD was originally designed for detection of arbitrary objects (including wide and tall objects). In the meantime, face bounding boxes are typically square. The authors propose to remove SSD features irrelevant for face detection. Finally, NMS algorithm (also used in MTCNN) has been improved. It has been noticed that when applied to videos, NMS results in visible face bounding box jitter between adjacent frames. The jitter might reduce applicability of face detection algorithm on videos. To resolve the problem a small regression module has been proposed instead of NMS to stabilize bounding box prediction. Similarly to MTCNN, BlazeFace also predicts face landmarks for each face detected (eyes, ears, mouth center, nose tip). Despite the improvements proposed, the authors haven't compared the work with other algorithms in the field. Also, neural network training has been performed on a closed dataset. In our work, we will compare the algorithm both in terms of inference speed and quality on several scenes. In contrast to the above-described algorithms, BlazeFace doesn't work with images of arbitrary resolution. BlazeFace has 2 modifications: for photos taken on front-facing camera with resolution of \(128 \times 128\) and on rear-facing camera at \(256 \times 256\). The authors expect photos taken on frontal camera to be quite large, and on rear camera to be smaller.

\section{Experiment Setup}

\begin{figure*}
  \centering
  \begin{subfigure}{0.49\linewidth}
    \centering
    \includegraphics[width=0.5\linewidth]{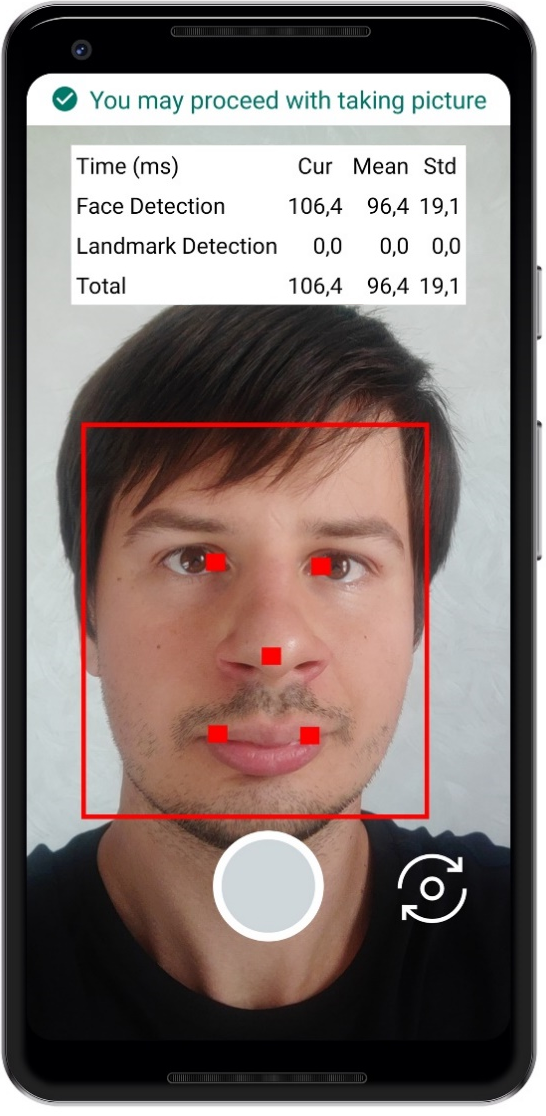}
    \caption{Face and landmark detection}
    \label{fig:mobile-app-inference}
  \end{subfigure}
  \begin{subfigure}{0.49\linewidth}
    \centering
    \includegraphics[width=0.5\linewidth]{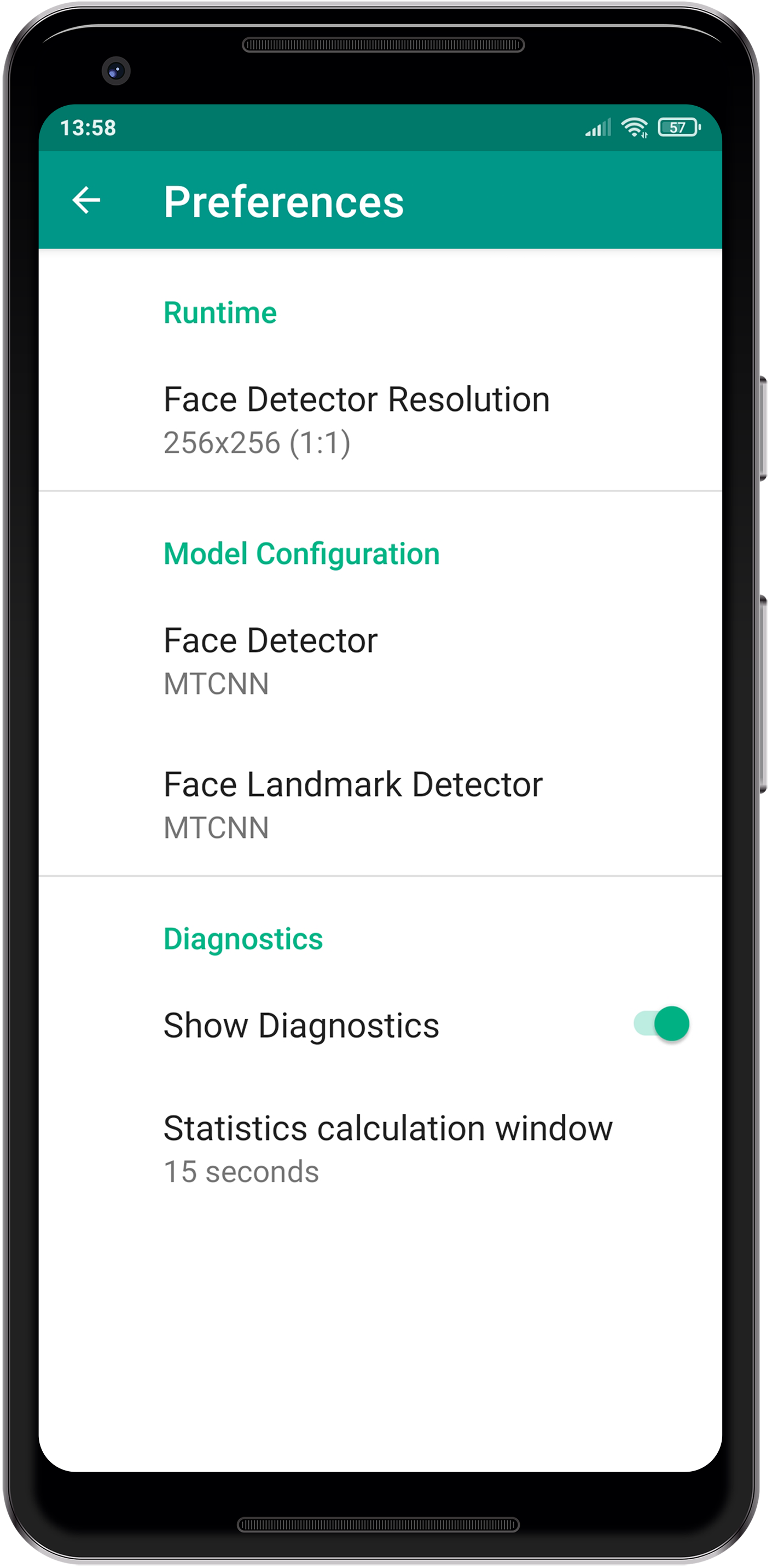}
    \caption{Configuration}
    \label{fig:mobile-app-config}
  \end{subfigure}
  \caption{Android face detection application interface. (a)~camera preview with inference time diagnostics (last value, mean and standard deviation). (b)~configuration, including input image resolution, model and diagnostics configuration.}\label{fig:mobile-app}
\end{figure*}

To conduct the experiments we have developed a mobile face detection application for Android OS. Face and landmark detection is shown in \cref{fig:mobile-app-inference}. The application has several configuration options available (\cref{fig:mobile-app-config}). First is algorithm input image resolution. There are typical values, such as \(128 \times 128\), \(256 \times 256\), \(480 \times 360\), \(640 \times 480\), and extremely low: \(32 \times 32\), \(64 \times 64\). The latter are used to test, whether certain algorithm is capable of face detection on images with low level of detail. Next, it is possible to configure face detection model. Here we can enable face landmark detection if the algorithm supports it. Finally, diagnostics configuration is available. An example of diagnostics information is available in \cref{fig:mobile-app-inference} on top, where information about last frame detection time, mean and standard deviation for inference times over a certain calculation window are presented.

We have selected 2 different smartphones with Android OS and Qualcomm Snapdragon processors. Smartphones with these processors are widely available on the market. Thus, our results will be easily reproducible. The first processor in Snapdragon 800, that is a flagship processor from 2013. Given year-over-year processor performance improvements, in 2022 its performance is similar to low-end processors. The second processor is Snapdragon 845, which is a flagship processor from 2018. In 2022 it can still be considered as mid to high-end CPU.

Next, each of the algorithms presented will be evaluated based on 3 criteria:
\begin{enumerate*}[label={\arabic*)}]
  \item inference time on a mobile device given certain configuration and CPU;
  \item whether the algorithm can detect face in one of the test scenarios;
  \item visual traits of the algorithms, such as bounding box jitter on video.
\end{enumerate*}

\section{Experiments}

The goal of the first experiment conducted is to estimate mean and standard deviation of algorithm inference time on both processors. In \cref{tab:inference-time} experiment results are shown for images of size \(256 \times 256\). Statistics are calculated over a timeframe of 30 seconds, with equal time devoted for scenes without faces, with single large face, and 2 small faces. The algorithms execute 2.13 -- 5.64 times slower on older Snapdragon 800 processor. Interestingly, neural-network-based algorithms have seen smaller relative slowdown, than algorithms with handcrafted features. Because-of this, Haar cascades algorithm has become slower to execute on Snapdragon 800 than both MTCNN and BlazeFace, while it was faster on Snapdragon 845.

\begin{table}
  \caption{Algorithm inference time comparison on mobile devices. \(\mu\) is mean, sd is sample standard deviation. Statistics are calculated over a timeframe of 30 seconds, with equal time devoted for scenes without faces, with single large face, and with 2 small faces. Image resolution is \(256 \times 256\). BlazeFace rear modification is used. Rightmost colum shows relative algorithm speedup on Snapdragon 845 with respect to Snapdragon 800.}\label{tab:inference-time}
  \centering
  \begin{tabular}{@{}lrrrrc@{}}
    \toprule
    Algorithm        & \multicolumn{2}{c}{Snapdragon 845} & \multicolumn{2}{c}{Snapdragon 800} & Speedup   \\
                     & \(\mu\) (ms) & sd                  & \(\mu\) (ms) & sd                  & (times)  \\
    \midrule
    MTCNN            & 98.9         & 24.1                & 210.3        & 100.4               & 2.13     \\
    BlazeFace        & 83.0         & 18.8                & 253.2        &  57.4               & 3.05     \\
    LBP              & 14.5         &  4.2                &  72.8        &  19.0               & 5.02     \\
    HAAR             & 71.2         &  7.2                & 401.8        &  87.6               & 5.64     \\
    HOG              & 25.2         &  4.0                &  90.6        &  28.1               & 3.60     \\
    \bottomrule
  \end{tabular}
\end{table}

\begin{figure*}
  \centering
  \begin{subfigure}{0.49\linewidth}
    \includegraphics[width=\linewidth]{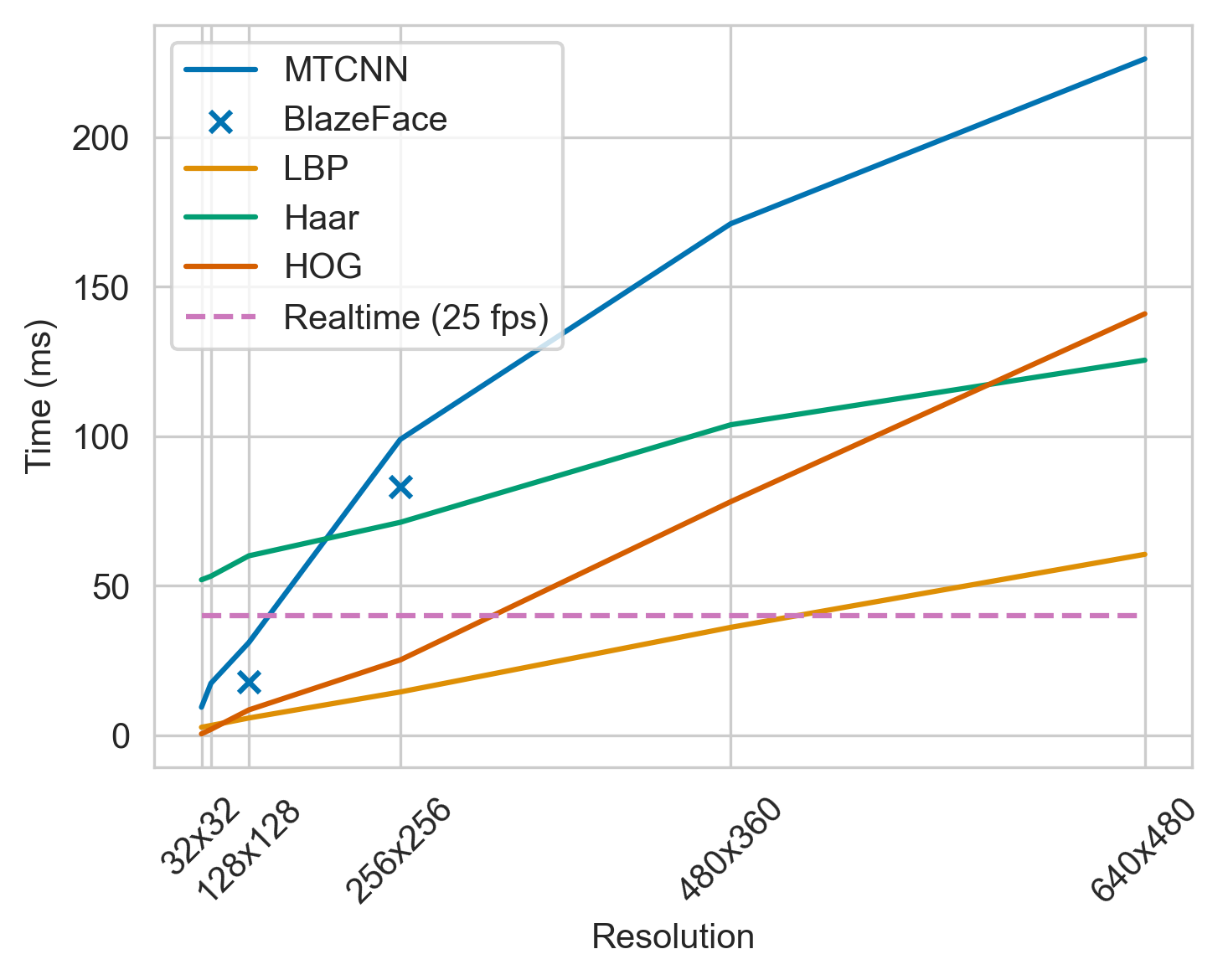}
    \caption{Snapdragon 845}\label{fig:inference-time-by-resolution-sd845}
  \end{subfigure}
  \begin{subfigure}{0.49\linewidth}
    \includegraphics[width=\linewidth]{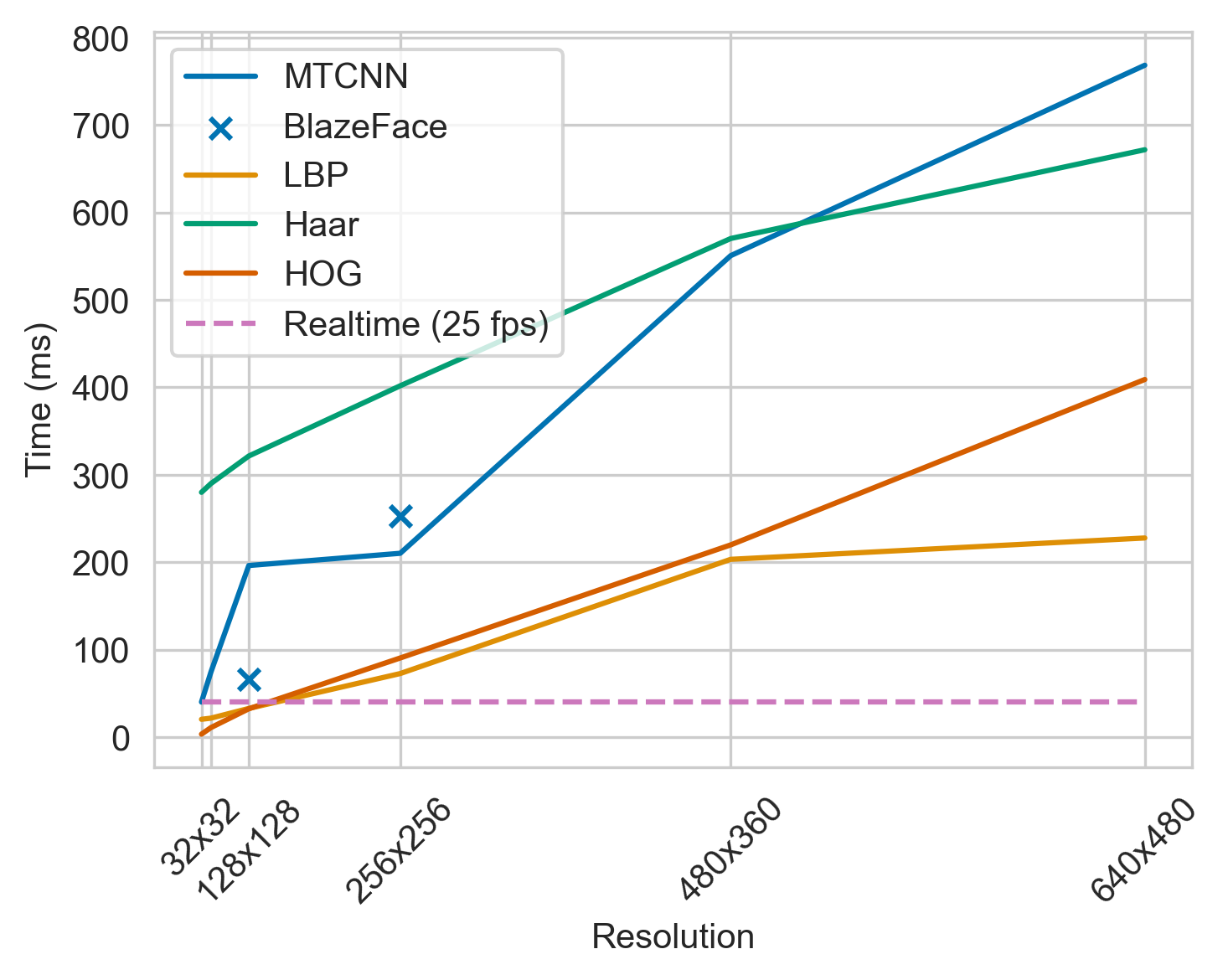}
    \caption{Snapdragon 800}\label{fig:inference-time-by-resolution-sd800}
  \end{subfigure}
  \caption{Face detection inference time comparison with respect to input image resolution. Most algorithms accept input images of arbitrary resolutions, while BlazeFace has two fixed modifications for front (\(128 \times 128\)) and rear (\(256 \times 256\)) cameras. Note, that on Snapdragon 845 all algorithms except Haar cascades can perform in realtime at resolution of \(128 \times 128\). LBP and HOG can even perform at \(256 \times 256\). Performance is drastically lower on Snapdragon 800. Only LBP and HOG algorithms show good performance at resolution of \(128 \times 128\) or below.}\label{fig:inference-time-by-resolution}
\end{figure*}

It has been noted previously, that most algorithms in this paper can work with input images of different resolutions. In \cref{fig:inference-time-by-resolution-sd845,fig:inference-time-by-resolution-sd800} we show inference time with respect to input resolution on Snapdragon 800 and 845 processors correspondingly. Dashed horizontal line depicts realtime inference baseline (at 25 frames per second or 40 milliseconds). Note, that BlazeFace has models for images taken on front-facing camera (at resolution of \(128 \times 128\)) and on rear-facing (at resolution of \(256 \times 256\)). As these models have somewhat different architectures, the results between these resolutions cannot be interpolated. Thus, BlazeFace results are shown with crosses.

Based on \cref{fig:inference-time-by-resolution} it is possible to conclude that Haar cascades (Viola-Jones) algorithm is quite slow in all testing scenarios. Other algorithms work in realtime on newer Snapdragon 845 at resolution of \(128 \times 128\). LBP and HOG algorithms can work fast-enough at resolution of \(256 \times 256\). In contrast, Snapdragon 800 shows poor performance on most algorithms. Only LBP, HOG execute in realtime at resolution of \(128 \times 128\) or below. Inference of BlazeFace (front) takes 65.9 ms (15.2 fps), which also might be acceptable in certain cases.

\begin{figure}
  \centering
  \includegraphics[width=0.95\linewidth]{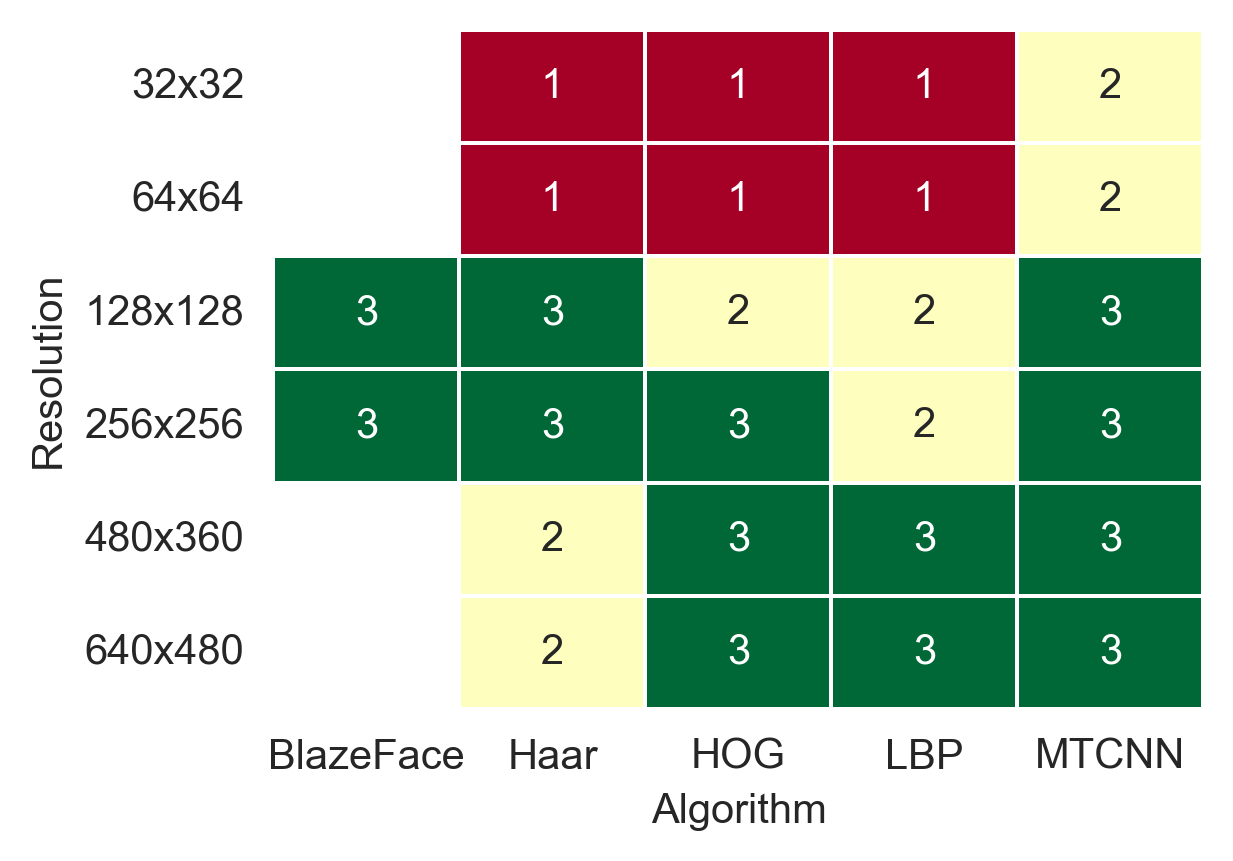}
  \caption{Visual algorithm evaluation on test scenarios. Algorithms that met all criteria are shown in green (score: 3). Algorithms with score of 2 and 1 are shown in yellow and red correspondingly. Red means that algorithm is inapplicable at this resolution. Note, face detection is impossible at resolutions of \(32 \times 32\) and \(64 \times 64\) with Haar, HOG, LBP methods. In the meantime MTCNN can still detect single large face.}\label{fig:visual-algorithm-testing}
\end{figure}

As it has been shown, not all algorithms can work fast-enough with images of large size. However, up to this point we didn't specify if there is any difference in face detection quality between different resolutions. As algorithms presented in this survey have been trained on different datasets, we will conduct visual quality testing targeting mobile face control system~\cite{RfidAccessControl}. During the experiment, we will score each algorithm from 0 to 3. 1 point is given for each of the following criteria:
\begin{enumerate*}[label={\arabic*)}]
  \item given a picture without any faces, the algorithm does not have false face detections;
  \item given a single large face on a picture, the algorithms finds it;
  \item given two small faces, the algorithm finds both of them.
\end{enumerate*}
Testing results are shown in \cref{fig:visual-algorithm-testing}. Cases with score of 1, 2, 3 are shown in red, yellow and green colors correspondingly. Note, that in our testing each algorithm has got at least 1 point.

Visual issues observed. \textbf{Haar Cascades:} on images of size \(480 \times 360\) or above, the algorithm finds non-existing faces. On a video, such face detection sporadically appears and disappears. \textbf{LBP:} on images of size \(256 \times 256\) and below the algorithm fails to find small faces. At \(64 \times 64\) or below no faces are detected. \textbf{HOG:} is quite good at finding faces on images at resolution of \(256 \times 256\) or above. \textbf{MTCNN:} is able to find single large face on photos even at resolution of \(32 \times 32\), which makes this algorithm the most adaptive among the selected algorithms. \textbf{BlazeFace:} has 2 fixed resolutions and works perfectly on both. It should be noted, that thanks to the special regression module, face bounding box predictions are much more stable on video in comparison to other algorithms.

\begin{figure*}
  \centering
  \begin{subfigure}{0.33\linewidth}
    \includegraphics[width=\linewidth]{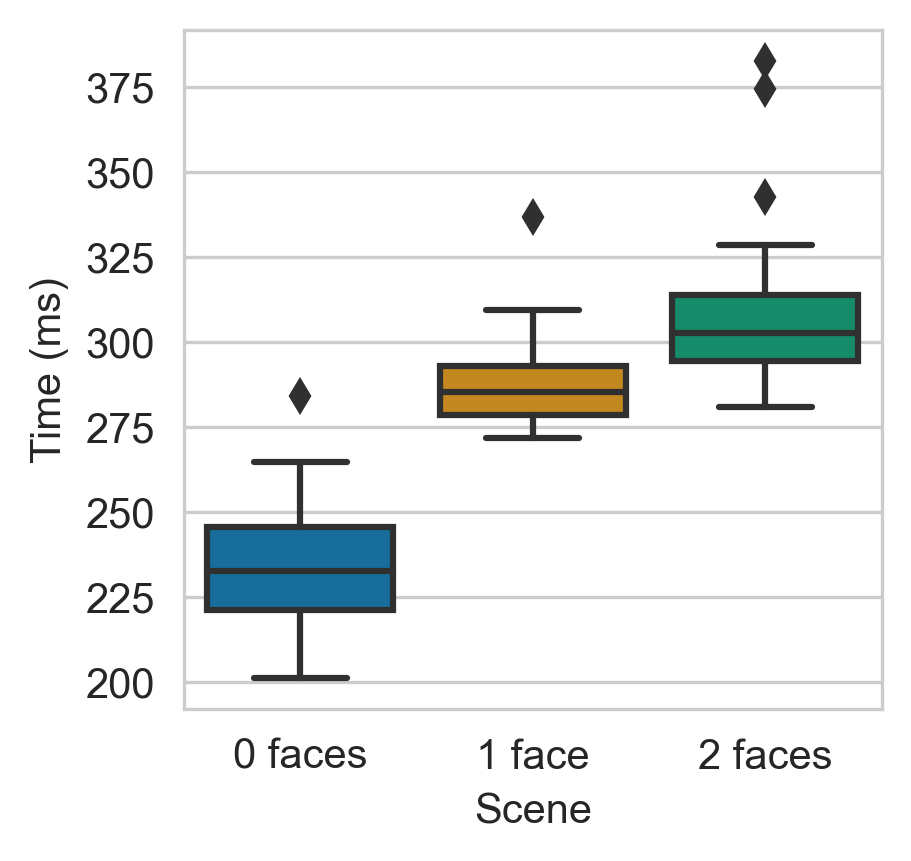}
    \caption{MTCNN}
  \end{subfigure}
  \begin{subfigure}{0.33\linewidth}
    \includegraphics[width=\linewidth]{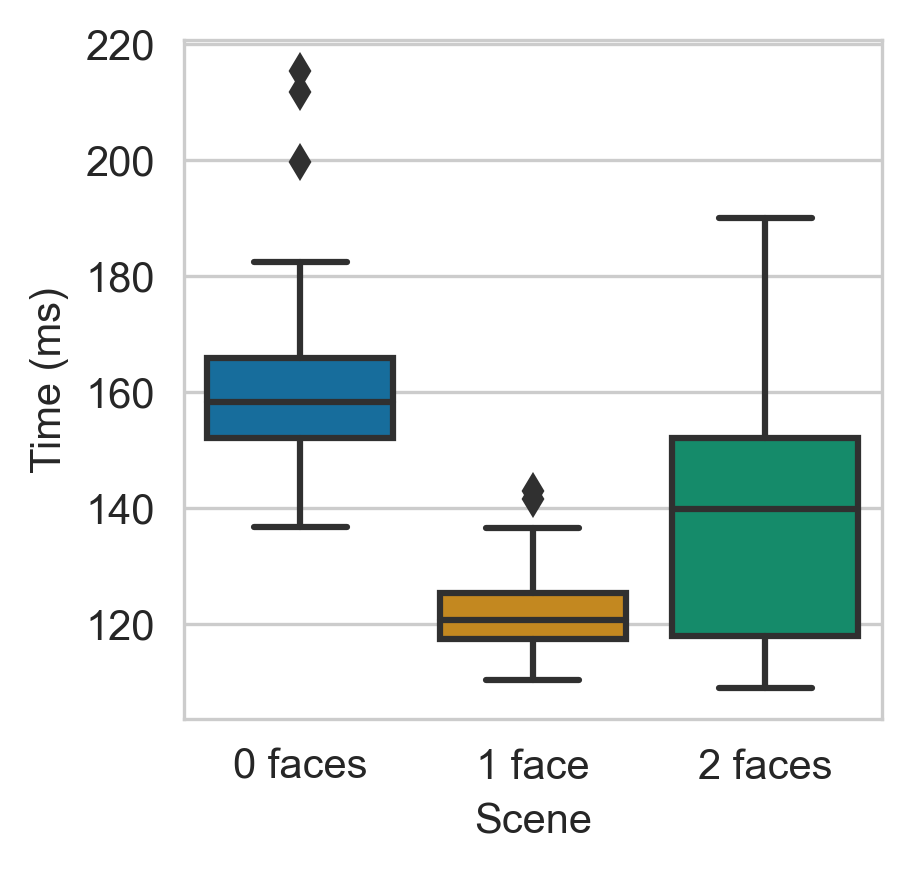}
    \caption{BlazeFace (rear)}
  \end{subfigure}
  \begin{subfigure}{0.33\linewidth}
    \includegraphics[width=\linewidth]{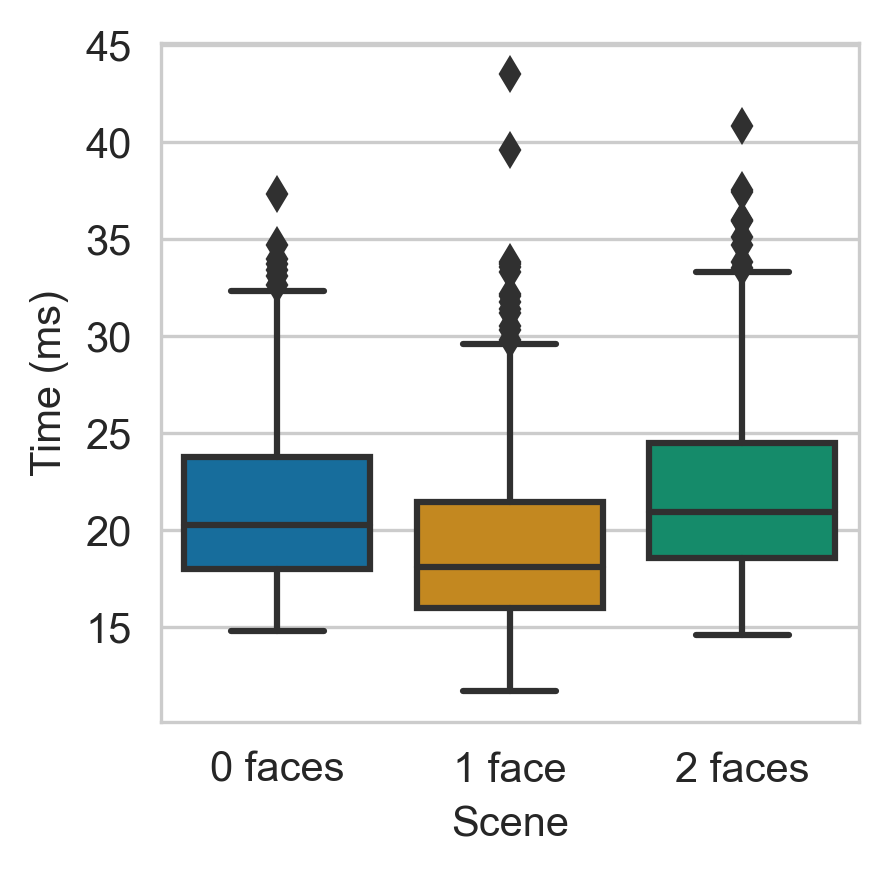}
    \caption{BlazeFace (front)}
  \end{subfigure}
  \begin{subfigure}{0.33\linewidth}
    \includegraphics[width=\linewidth]{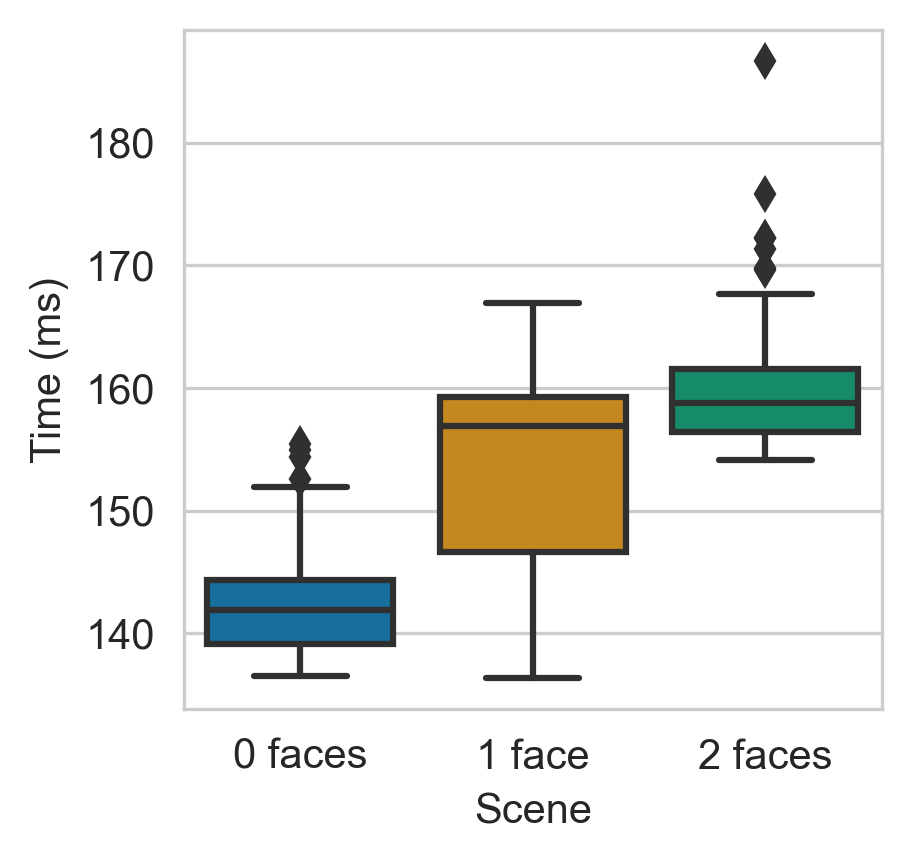}
    \caption{HOG}
  \end{subfigure}
  \begin{subfigure}{0.33\linewidth}
    \includegraphics[width=\linewidth]{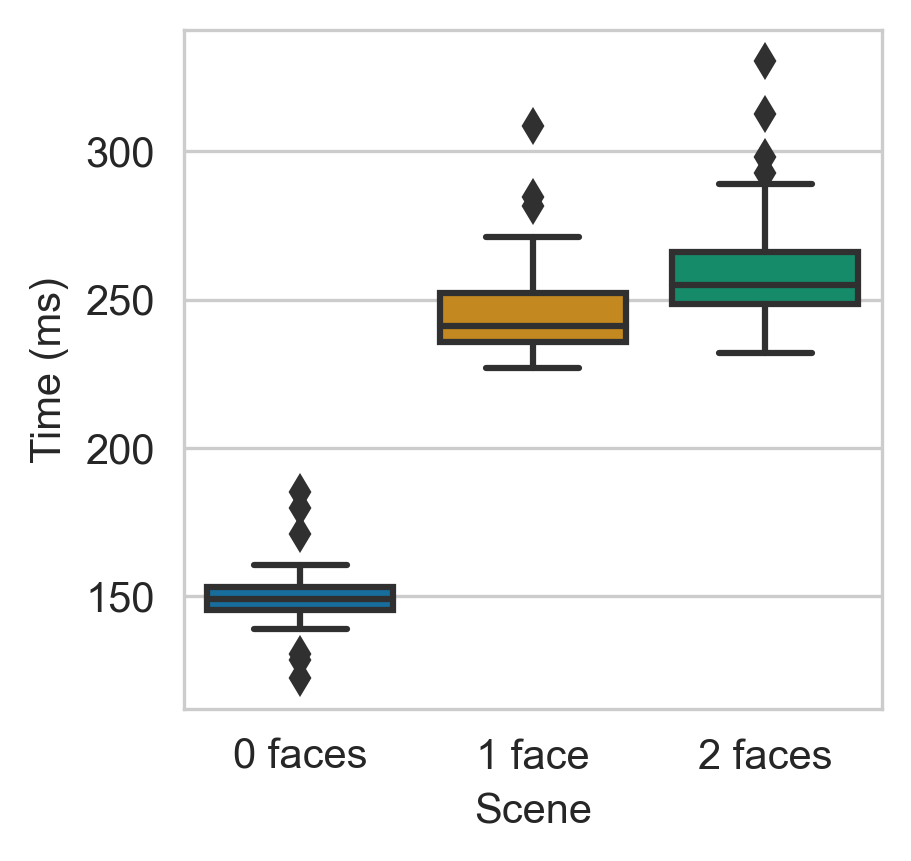}
    \caption{Haar}
  \end{subfigure}
  \begin{subfigure}{0.33\linewidth}
    \includegraphics[width=\linewidth]{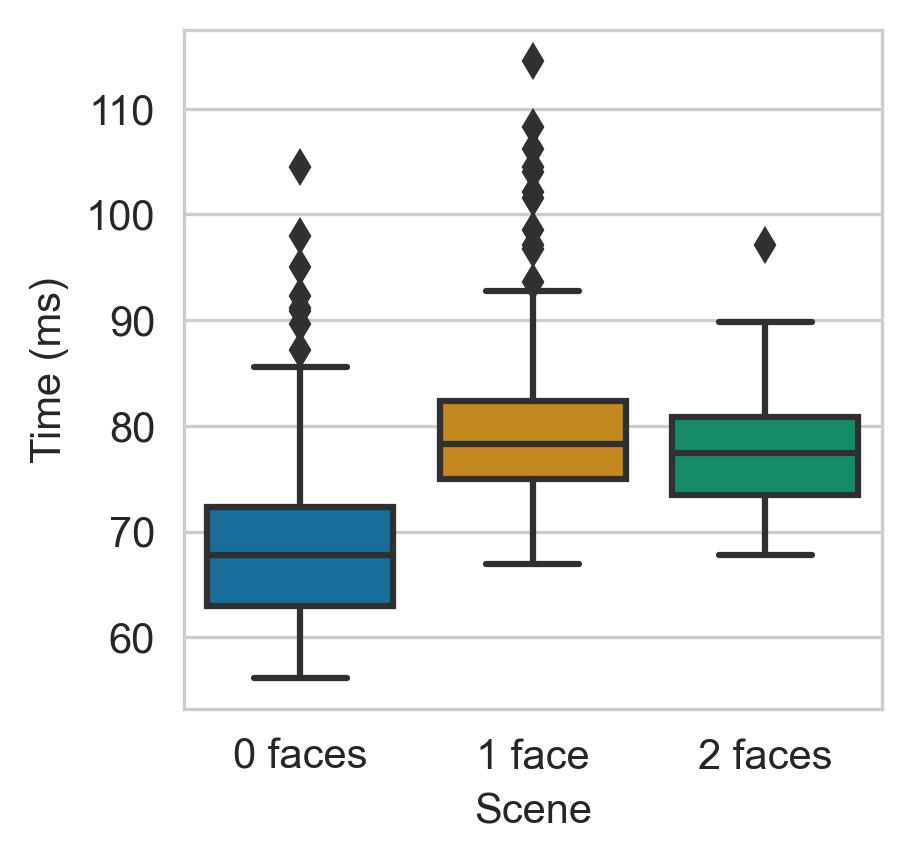}
    \caption{LBP}
  \end{subfigure}
  \caption{Inference time box plots for face detection algorithms. Measurements taken over 15s video on scenes with 0, 1, or 2 faces. Image resolution is set to \(640 \times 480\) for all algorithms, except BlazeFace~(rear) at \(256 \times 256\) and BlazeFace~(front) at \(128 \times 128\). Note, that inference time of most algorithms clearly growth with the number of faces present of the image. Interestingly, BlazeFace~(rear) shows maximum inference time when no faces are present. BlazeFace~(front) shows no clear trend.}\label{fig:inference-time-by-face-number}
\end{figure*}

Based on the data from \cref{tab:inference-time} it has been noticed, that standard deviation of the algorithms is substantially different. It is interesting to understand the roots of such behavior. In \cref{fig:inference-time-by-face-number} we show box plots for inference times.
Statistics were calculated over a timeframe of 15 seconds. The measurements are split into testing scenes with 0, 1, and 2 faces. Most algorithms clearly show inference time growth with the number of faces present in the image. For MTCNN, Haar, LBP algorithms this can be explained by their cascaded nature, where the fewer image regions contain face-like objects, the earlier such regions are rejected. Interestingly, BlazeFace~(rear) shows maximum inference time when no faces are present. BlazeFace~(front) shows no clear trend. Such behavior is less efficient for cases, when most frames contain no faces.

\section{Conclusions}

Five face detection algorithms have been adapted to smartphones in this work. Based on the conducted analysis, it is possible to conclude that the best algorithms in terms of mobile applicability are:
\begin{itemize}
  \item BlazeFace, whose face bounding box predictions are stable (have almost no jitter) in videos. However, the algorithm accepts input images of only 2 fixed resolutions, which is a drawback. Also, high inference time on images without faces is untypical;
  \item MTCNN, which thanks to the cascaded architecture is able to conserve computational resources on frames without faces. It is the algorithm that can detect faces on the widest range of input resolutions;
  \item if inference time is the most important, face detector based on Histogram of Oriented Gradients (HOG) can be used. This algorithm has higher speed at a cost of some quality.
\end{itemize}
Also, in the work it has been shown that cascaded architectures adaptively change inference time based on input image contents (\ie whether there are any patches similar to a face).

We hope that the analysis and practical recommendations presented in this work will widen applicability of the described face detection methods in mobile applications.

\begin{small}
  \subsection*{Funding}

  The work is supported by the state budget scientific research project of Dnipro University of Technology ``Development of New Mobile Information Technologies for Person Identification and Object Classification in the Surrounding Environment'' (state registration number 0121U109787).

  \printbibliography
\end{small}

\end{document}